\pdfoutput=1

\documentclass[11pt]{article}

\usepackage[preprint]{acl}

\usepackage{times}
\usepackage{latexsym}

\usepackage[T1]{fontenc}

\usepackage[utf8]{inputenc}

\usepackage{microtype}

\usepackage{inconsolata}

\usepackage{graphicx}

\usepackage{CJKutf8}

\usepackage{tabularx}

\usepackage{booktabs}  
\usepackage{multirow}  
\usepackage{pgfplots}
\usepackage{siunitx}

\author{
    Ke-Ching Chang,\textsuperscript{\rm 1}
    Chung-Chi Chen,\textsuperscript{\rm 2}
    An-Zi Yen\textsuperscript{\rm 1}
    \\
    \textsuperscript{\rm 1} Department of Computer Science, National Yang Ming Chiao Tung University, Taiwan
    \\
    \textsuperscript{\rm 2} Artificial Intelligence Research Center, AIST Japan
    \\
    \texttt{cassie.cs09@nycu.edu.tw,}
    \texttt{c.c.chen@acm.org,}
     \texttt{azyen@nycu.edu.tw}
}

\title{Paraphrase-Aligned Machine Translation}

\begin{document}
\begin{CJK*}{UTF8}{bkai}
\maketitle
\begin{abstract}
Large Language Models (LLMs) have demonstrated significant capabilities in machine translation. However, their translation quality is sometimes questioned, as the generated outputs may deviate from expressions typically used by native speakers. These deviations often arise from differences in sentence structure between language systems. To address this issue, we propose ParaAlign Translator, a method that fine-tunes LLMs to paraphrase sentences, aligning their structures with those of the target language systems. This approach improves the performance of subsequent translations. Experimental results demonstrate that the proposed method enhances the LLaMA-3-8B model's performance in both resource-rich and low-resource scenarios and achieves parity with or surpassing the much larger LLaMA-3-70B model.
\end{abstract}

\section{Introduction}
Direct translations often result in unnatural or unidiomatic expressions that native speakers would not typically use. While most translators can produce semantically consistent translations, the output often lacks fluency or idiomaticity. For example, the Chinese phrase ``一般人'' is frequently translated as ``ordinary people'' in English. However, a native English speaker might prefer the phrase ``not famous enough'' to convey the same idea. This discrepancy suggests that rephrasing the original Chinese sentence to better align with the structure and expressions of the target language could enhance the translation. If ``一般人 (ordinary people)'' were rewritten in Chinese as ``不夠 (not enough) 有名的人 (famous person),'' a more explicit phrase, a translator would be more likely to render it into the idiomatic English phrase ``not famous enough.'' Therefore, improving the fluency and naturalness of translation outputs through reformatting the source sentence before translation is a valuable area of exploration.

Recently, Large Language Models (LLMs) have demonstrated remarkable capabilities across a broad range of natural language processing (NLP) tasks, including machine translation. Nonetheless, translated texts generated by LLMs often sound unnatural and do not reflect typical expressions used by native speakers. 
Taking Chinese-to-English translation as an example, if the target English sentence is ``He is not famous enough,'' LLMs can accurately produce this translation when given the source Chinese sentence ``他是不夠有名的人。'' However, the alternative Chinese sentence ``他是一般人,'' which conveys the same meaning, cannot be translated into the target sentence. 
To address this challenge, we introduce ParaAlign Translator, designed to enhance the translation quality of LLMs by leveraging instruction tuning to guide them in paraphrasing the original sentences prior to translation. This approach generates rephrased sentences that are grammatically and semantically aligned with the target language, thereby improving overall translation effectiveness. Our experiments cover both resource-rich and low-resource languages, with results demonstrating that this method can effectively enhance performance across diverse scenarios.

The primary contributions of this work are as follows:
(1) We propose ParaAlign Translator, a novel method that strengthens the translation capabilities of LLMs by fine-tuning them with aligned pairs of original and paraphrased sentences.
(2) We demonstrate that ParaAlign Translator outperforms LLMs employing few-shot prompting, even when compared with larger model sizes. Moreover, we observed significant performance improvements using our approach with only 5\% of the original dataset.
(3) We conduct extensive evaluations across a range of languages, including both high-resource and low-resource settings. Our experiments encompass translations between English and Chinese, English and German, English and Hebrew, and English and Swahili.

\section{Related Work}

\citet{wu-etal-2023-wspalign} proposed contrastive alignment instructions (AlignInstruct) to enhance the machine translation (MT) performance of large language models (LLMs). Building on the concept of alignment, \citet{puduppully-etal-2023-decomt} introduced the DecoMT approach, which further improves the translation capabilities of LLMs, particularly for closely related languages. Utilizing few-shot prompting, this method simplifies the translation process by breaking it down into word chunk translations, thereby significantly improving both fluency and accuracy. They posited that their method achieves higher performance by reordering sub-sentences, thus alleviating the burden on LLMs. In other words, when the sentence structures of the source and target languages are similar, LLMs can produce higher-quality translations. Inspired by these works, we extend this approach by fine-tuning models with additional paraphrased data pairs to teach LLMs how to align sentences more effectively by reordering the source language sentence structure to better match that of the target language.

Few-shot prompting uses a few examples to guide LLMs for consistent responses. 
This approach facilitates the generation of satisfactory responses without requiring model fine-tuning. Numerous studies \citep{NEURIPS2020_1457c0d6, zhang-etal-2023-machine} have demonstrated that LLMs guided by few-shot prompting can perform well in machine translation tasks. Moreover, \citet{pmlr-v202-zhang23m} observed that increasing the number of prompt examples significantly enhances the average translation quality of LLMs. Additionally, \citet{cahyawijaya-etal-2024-llms} demonstrated that few-shot learning can improve the machine translation capabilities of LLMs, even for low-resource languages. Consequently, when generating paraphrased language pairs, we employ few-shot prompting to guide LLMs effectively.

\begin{table} [t]
  \centering
  \small
  \begin{tabular}{l p{6.5cm}}
    \hline
    \textbf{ID} & \textbf{Prompt} \\
    \hline
    \textbf{P1}     & {Please translate the following sentence from \verb|SRC| to \verb|TGT|. 
    
    Here is some examples: 

    \#\#\#\verb|SRC|: \textbf{S1}
    
    \#\#\#\verb|TGT|: \textbf{S2}
    
    \#\#\#\verb|SRC|: \textbf{S3}
    
    \#\#\#\verb|TGT|: \textbf{S4}}           \\
    \textbf{P2}     & {Please translate the following sentence from \verb|SRC| to \verb|TGT|.}           \\
    \textbf{P3}     & {Convert the following \verb|SRC| sentence into another \verb|SRC| sentence that maintains the same meaning but is more likely to translate into natural, native-sounding \verb|TGT|.}           \\
  \hline
  \end{tabular}
  \caption{\label{tab:first_table}The prompts used in this paper serve distinct purposes: P1 is applied to direct translation tasks, while P3 is utilized for paraphrasing tasks and works in conjunction with P2 during model training. Here, \texttt{SRC} and \texttt{TGT} denote the source and target languages, respectively, and S represents a sentence.}
\end{table}

\begin{table*}[t]
    \resizebox{\textwidth}{!}{%
\begin{tabular}{l|cccccccc}
 & \multicolumn{2}{c}{\textbf{Zh $\Rightarrow$ En}} & \multicolumn{2}{c}{\textbf{En $\Rightarrow$ Zh}} & \multicolumn{2}{c}{\textbf{De $\Rightarrow$ En}} & \multicolumn{2}{c}{\textbf{En $\Rightarrow$ De}} \\
                        & \text{COMET} & \text{ROUGE-L} & \text{COMET} & \text{ROUGE-L} & \text{COMET} & \text{ROUGE-L} & \text{COMET} & \text{ROUGE-L} \\
\hline
\text{LLaMA-3-8B}          & 79.65 & 47.85 & 82.34 & 17.81 & 81.94 & 55.16 & 82.85 & 47.87 \\
\text{ParaAlign Translator}     & \textbf{79.90} & \textbf{50.67} & \textbf{83.52} & \textbf{18.70} & \textbf{83.69} & \textbf{59.44} & \textbf{83.11} & \textbf{49.52} \\
\end{tabular}
}
\caption{Results of resource-rich languages.}
\label{tab:Results of resource-rich languages.}
\end{table*}

\begin{table*}[t]
    \resizebox{\textwidth}{!}{%
\begin{tabular}{l|cccccccc}
& \multicolumn{2}{c}{\textbf{Heb $\Rightarrow$ En}} & \multicolumn{2}{c}{\textbf{En $\Rightarrow$ Heb}} & \multicolumn{2}{c}{\textbf{Swh $\Rightarrow$ En}} & \multicolumn{2}{c}{\textbf{En $\Rightarrow$ Swh}} \\
                        & \text{COMET} & \text{ROUGE-L} & \text{COMET} & \text{ROUGE-L} & \text{COMET} & \text{ROUGE-L} & \text{COMET} & \text{ROUGE-L} \\
\hline
\text{LLaMA-3-8B}          & 83.95 & 58.45 & 83.04 & \textbf{24.13} & 81.37 & 55.94 & \textbf{77.68} & \textbf{39.35} \\
\text{ParaAlign Translator}     & \textbf{86.66} & \textbf{65.56} & \textbf{84.80} & 22.51 & \textbf{84.18} & \textbf{63.02} & 71.64 & 35.62 \\
\end{tabular}
}
\caption{Results of low-resource languages.}
\label{tab:Results of low resource languages.}
\end{table*}

\section{Method}
The proposed ParaAlign Translator consists of two phases.
The first phase involves collecting bilingual sentence pairs and utilizing LLMs to generate back-translations. The prompt used for this task, P1, is shown in Table~\ref{tab:first_table}. These original sentences and their back-translated versions are subsequently employed to train a model in the second stage, enabling it to rephrase the original sentences with improved wording and structures for better translation. 
We use \texttt{LLaMA-3-8B} \cite{llama3modelcard} to generate paraphrased and aligned sentence pairs as training data for subsequent model fine-tuning. For instance, to collect Chinese (Zh) and paraphrased Chinese (Zh') sentence pairs, we apply few-shot prompting, using a English dataset as input to generate the corresponding paraphrased Chinese sentences (Zh').

In the second phase, the \texttt{LLaMA-3-8B} model is fine-tuned using Low-Rank Adaptation (LoRA) \citep{hu2022lora}, which significantly reduces the number of trainable parameters. Our approach leverages both target-to-source language pairs and target-to-paraphrased language pairs as training data. Additionally, Prompts P2 and P3, described in Table~\ref{tab:first_table}, are employed in this phase. By incorporating additional paraphrased-aligned sentence pairs into the training data, the model acquires the capability to first paraphrase and align source sentences, enabling their structure to more closely match the syntax of the target language.

\section{Experiment}
\subsection{Experimental Setup}
In this paper, the training and testing datasets are sourced from \citet{wang-etal-2024-taste}. Specifically, the WMT validation set and MTME multi-candidate dataset are utilized to construct the training data, while the WMT22 test set is employed as the testing data. The ParaAlign Translator is evaluated on translations between English (En) and Chinese (Zh), as well as English (En) and German (De). The training dataset contains 21,966 sentences, and the test dataset comprises 3,859 sentences. 

Additionally, for low-resource languages, Hebrew (Heb) and Swahili (Swh), data is sourced from the FLORES-200 dataset \citep{nllb-24}. Since the FLORES-200 test set is not publicly accessible, the dev-test set is used as the test set. A total of 505 sentences are selected from the dev-test set for testing, while the remaining 1,502 sentences are utilized as the training set. 

The backbone model is \texttt{LLaMA-3-8B}, fine-tuned using LoRA \citep{han2023unsloth}. For the LoRA hyperparameters, we adhere to the default settings, with the exception of setting the LoRA rank to 128. 
The temperature is set to 0.001. As baseline comparisons, we use the \texttt{LLaMA-3-8B} and \texttt{LLaMA-3-70B} \cite{llama3modelcard} models without fine-tuning.
For evaluation, we adopt ROUGE-L~\citep{lin-2004-rouge} and COMET scores using \texttt{wmt22-comet-da} ~\citep{rei-etal-2022-comet}.

\subsection{Experimental Result}
The results for resource-rich languages are presented in Table~\ref{tab:Results of resource-rich languages.}. These results demonstrate that the proposed ParaAlign Translator consistently outperforms \texttt{LLaMA-3-8B} across all language pairs and metrics. Additionally, the approach was evaluated on low-resource languages, specifically Hebrew and Swahili. As shown in Table~\ref{tab:Results of low resource languages.}, the ParaAlign Translator achieves a 2.71\% improvement in the COMET score and a 7.11\% increase in the ROUGE-L score compared to the baseline performance of \texttt{LLaMA-3-8B}. These findings underscore the effectiveness of the proposed approach in enhancing translation performance for low-resource languages. Overall, the experimental results confirm that the ParaAlign Translator produce high-quality translations by jointly learning to paraphrase and translate. This demonstrates that fine-tuning the model to paraphrase input sentences prior to translation is a promising strategy for achieving superior translation quality.

\begin{table}[t]
  \centering
  \small
  \begin{tabular}{l p{5cm}}
    \hline
    \textbf{Method} & \textbf{Prompt} \\
    \hline
    \textbf{SRC}     & {以免再次发生这样的事情}           \\
    \textbf{TGT}     & {So that such a thing won’t happen again.}           \\
    \textbf{LLaMA-3-8B}     & {To prevent such incidents from happening again.}           \\
    \textbf{PT (Our)}     & {So that it doesn't happen again.}           \\
    \hline
  \end{tabular}
  \caption{Case study of Zh $\Rightarrow$ En. PT denotes ParaAlign Translator.}
  \label{tab:forth_table}
\end{table}

\section{Further Analysis}
In this section, we present a case study, conduct various model comparisons, and analyze the impact of training data size using the (Zh $\Rightarrow$ En) dataset.

\subsection{Case Study}
In Table~\ref{tab:forth_table}, we compare translations generated by the ParaAlign Translator with those produced by the \texttt{LLaMA-3-8B} model, which has been fine-tuned only on original sentence pairs (Zh $\Rightarrow$ En). Traditional approaches, while often generating grammatically correct translations, tend to overlook inherent structural differences between Chinese and English, such as variations in word order and sentence construction. Consequently, translations from such models may feel unnatural or lack fluency, as observed in the output from the \texttt{LLaMA-3-8B} model.

In contrast, the ParaAlign Translator considers the structural nuances and contextual dependencies of both languages, enabling it to produce translations that are not only accurate but also aligned with natural English phrasing. This capability ensures that the translated output is closer to the ground truth, as demonstrated in the improved translation provided in the PT (ParaAlign Translator) row. The ParaAlign Translator consistently generates fluent and coherent translations.
For instance, the ParaAlign Translator's output, ``So that it doesn't happen again,'' is more natural and succinct than the \texttt{LLaMA-3-8B} model's translation, ``To prevent such incidents from happening again.'' While the latter is correct, it introduces unnecessary formality and slightly awkward phrasing. By better aligning the structural differences between the two languages, the ParaAlign Translator captures the intended meaning with a more conversational tone.

This case shows the strength of the ParaAlign Translator in managing structural differences and producing translations that closely align with native language usage. Such advantages significantly enhance the quality of machine-generated translations in real-world applications.

\begin{table}[t]
  \centering
  \small
    \begin{tabular}{ll|rr}
    \multicolumn{1}{c}{LLM} & \multicolumn{1}{c|}{Scheme} & \multicolumn{1}{c}{COMET} & \multicolumn{1}{c}{ROUGE-L} \\
    \hline
    \multirow{2}[1]{*}{LLaMA-3-8B} & Fine-Tuning & 79.11 & 47.29 \\
          & PT (Our) & 79.90 & \textbf{50.67} \\
    \hline
    LLaMA-3-70B & Few-Shot & \textbf{80.24} & 50.32 \\
    \end{tabular}%
   \caption{Variants of the model.}
  \label{tab:Variants of the model.}
\end{table}%

\subsection{Ablation and Model Size Comparison}
This section provides a comparison of different schemes for utilizing LLMs and evaluates the performance of an 8B model against a 70B model. First, to highlight the importance of fine-tuning with additional paraphrased sentences, we conducted an experiment using a training set that consisted only of original sentence pairs. Specifically, for the ParaAlign Translator targeting Chinese (Zh) $\Rightarrow$ English (En) translations, the training set included only Zh $\Rightarrow$ En pairs, excluding any Zh $\Rightarrow$ Zh' pairs. As shown in Table~\ref{tab:Variants of the model.}, the performance of the model fine-tuned exclusively with original sentence pairs is inferior to that of the model trained using our proposed method. This experiment underscores the critical role of our approach in enhancing the translation capabilities of LLMs.
Second, we conducted experiments with the LLaMA-3-70B model. As shown in Table~\ref{tab:Variants of the model.}, the proposed ParaAlign Translator (8B) achieves performance comparable to the significantly larger model when evaluated using the COMET metric and surpasses it when assessed with the ROUGE-L metric.

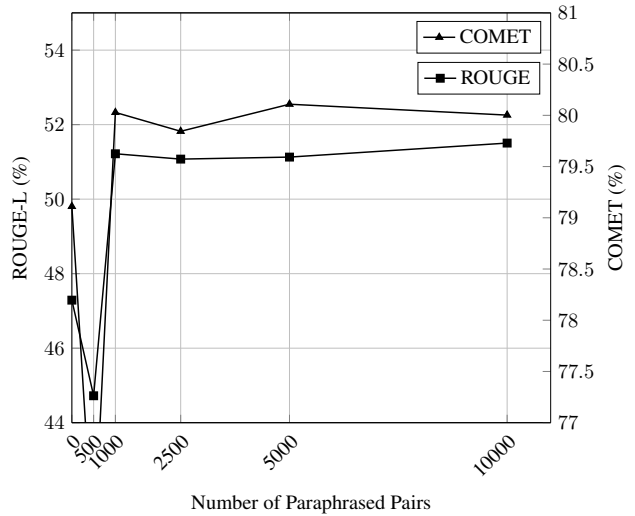
\begin{figure}[t]
\centering
\pgfplotsset{width=9cm,compat=1.14}
\begin{tikzpicture}[scale=0.7]
\begin{axis}[
  scale only axis,
  xlabel={Number of Paraphrased Pairs},
  ylabel={ROUGE-L (\%)},
  xmin=0, xmax=11000,
  ymin=44, ymax=55,
  xtick={0, 500, 1000, 2500, 5000, 10000},
  xticklabels={0, 500, 1000, 2500, 5000, 10000},
  xticklabel style={rotate=45, anchor=north east},
  legend style={at={(0.85,0.88)}, anchor=north, legend columns=2},
  grid=major,
  yticklabel style={black},
  axis y line*=left, 
  scaled x ticks=false, 
]

\addplot[mark=square*, color=black, thick]
    coordinates{
    (0,47.2892) (500,44.7229) (1000,51.2182) (2500,51.0748) (5000,51.1289) (10000,51.5051)
    };
\addlegendentry{ROUGE}

\end{axis}

\begin{axis}[
  scale only axis,
  ylabel={COMET (\%)},
  xmin=0, xmax=11000,
  ymin=77, ymax=81,
  xtick=\empty, 
  yticklabel style={black},
  axis y line*=right, 
]

\addplot[mark=triangle*, color=black, thick]
    coordinates{
    (0,79.1109) (500,75.7251) (1000,80.0284) (2500,79.8441) (5000,80.1084) (10000,80.0022)
    };
\addlegendentry{COMET}

\end{axis}
\end{tikzpicture}
\caption{Relationship between the number of paraphrased sentence pairs and performances.}
\label{fig:line_chart}
\end{figure}

\subsection{Data Size}

We further examine the relationship between the size of paraphrased sentence pairs and their corresponding scores. As illustrated in Figure~\ref{fig:line_chart}, it is notable that when the paraphrased data size is 500 (approximately 0.25\% of the original data size), the Rouge-L score is only 44.72\%, which is lower than the 47.29\% achieved using traditional training methods. However, when the paraphrased data size increases to 1,000 (around 5\% of the original data size), the Rouge-L score rises significantly to 51.22\% and remains relatively stable with further increases in data size. 
The COMET score also shows a similar trend.
These findings suggest that a minimal paraphrased data size may adversely affect model performance. In contrast, once the data size surpasses a certain threshold (approximately 5\% of the original data size in this case), the model is better equipped to learn the sentence structure of the target language, thereby improving its performance.

\section{Conclusion}

We propose ParaAlign Translator, a method that improves machine translation quality by fine-tuning LLMs with paraphrased sentence pairs. This approach enhances the fluency and idiomaticity of translations by aligning sentence structures between source and target languages. Our experiments show that ParaAlign Translator outperforms baseline models, including LLaMA-3-8B and LLaMA-3-70B, across multiple languages. Even with a small amount of paraphrased data (5\% of the original dataset), our method significantly boosts performance, highlighting paraphrasing as an effective strategy for improving machine translation quality.

\section*{Limitations}
Currently, the ParaAlign Translator has been tested primarily on English-to-other languages and other languages-to-English translation tasks. While this covers a broad range of scenarios, the method has not yet been tested on translation tasks involving pairs of non-English languages (i.e., language X to language Y, where both X and Y are non-English languages). The effectiveness of ParaAlign Translator in such cases remains uncertain, as the structural differences and paraphrasing challenges may differ when both source and target languages are non-English. Future work should explore this area to evaluate the method’s generalizability to non-English-to-non-English translations and determine whether additional adaptations are required.

\bibliography{custom}

\end{CJK*}
\end{document}